\documentclass{article} 
\usepackage{iclr2027_conference,times}
\usepackage{graphicx}
\usepackage{booktabs}

\usepackage{amsmath,amsfonts,bm}

\def\eqref#1{equation~\ref{#1}}

\def\1{\bm{1}}

\DeclareMathAlphabet{\mathsfit}{\encodingdefault}{\sfdefault}{m}{sl}
\SetMathAlphabet{\mathsfit}{bold}{\encodingdefault}{\sfdefault}{bx}{n}

\usepackage{hyperref}
\usepackage{url}

\title{Increasing Width Allows Greedy Layer-wise Training to Rival End-to-End Backpropagation in Self-Supervised Learning}

\author{Syon Mansur \& Joel Zylberberg  \\
Jules Stein Eye Institute\\
University of California, Los Angeles\\
Los Angeles, CA 90095, USA \\
\texttt{\{syonmansur, joelzy\}@ucla.edu} \\
}

\iclrfinalcopy
\begin{document}

\maketitle

\begin{abstract}

End-to-end backpropagation has been the dominant mode of training in deep learning, allowing for the coordination of parameter updates across layers of a neural network. Prior studies have explored alternative – and, in some cases, simpler – training mechanisms, showing that they can sometimes achieve performance similar to backpropagation. However, the architectural conditions under which locally optimized networks, which avoid end-to-end backpropagation of error, can learn representations comparable to those learned through end-to-end training remain unclear. We aim to answer this question in the context of self-supervised learning, an important framework for large-scale pretraining in artificial intelligence. Here, we investigate how network width and depth affect the efficacy of greedy layer-wise and end-to-end self-supervised training in convolutional networks. We find that in wider networks, the benefits of end-to-end backpropagation over greedy layer-wise training shrink: in relatively shallow and very wide networks, we even observed higher performance in models trained with greedy layer-wise training. Subsequent analysis of the representations formed by these networks shows that very wide greedy-trained networks exhibit more favorable representational geometry than do networks trained end-to-end with backpropagation. This work shows that width can compensate for restricted credit assignment and identifies differences in representational geometry as a potential mechanism for their improved performance.

\end{abstract}

\section{Introduction}

End-to-end backpropagation is the predominant training paradigm used in deep learning and has been responsible for much of the progress in modern representation learning \citep{lillicrap2020Backpropagation, lecun2015Deep}. End-to-end training propagates error signals throughout the entire network, allowing parameters across the network to be adjusted according to their impact on the network’s output \citep{rumelhart1986Learning}. While effective, end-to-end training requires error signals to travel across the entire network and requires retaining the intermediate activations for each backward pass. This can result in limitations in particularly deep or memory-constrained networks \citep{chen2016Training, lowe2020Putting, mostafa2018Deep}. Local learning, in contrast, restricts optimization to individual layers or modules, allowing for modular training or learning on hardware where global communication is costly. Previous work has demonstrated that local learning rules can be competitive with end-to-end training \citep{xiong2020LoCo, wang2021Revisiting, nokland2019Training, belilovsky2020Decoupleda, belilovsky2019Greedy}. Nevertheless, local learning often underperforms end-to-end training and the existing successes vary greatly in architectures and loss functions \citep{illing2021Locala}. Thus, it remains unclear when local learning can or cannot create high quality representations. We address this knowledge gap by comparing greedy layer-wise training to end-to-end training for models of varying widths and depths in the context of self-supervised learning.

Greedy layer-wise learning is a local learning paradigm that optimizes each layer or stage of a network sequentially rather than propagating an error signal throughout the entire network. In greedy layer-wise training, after a layer is trained, its parameters are frozen, its outputs are provided as inputs to the next layer, the next layer is trained, frozen, and so on. This can be simpler to implement than end-to-end backpropagation, but it comes at a cost: information about the model's output layers is not available to inform parameter updates at earlier layers. In this study, we identify architecture choices that can reduce this cost, to determine when greedy layer-wise training can yield comparable performance to backpropagation.

Theoretical studies of expressivity, the ability of a neural network to represent highly complex functions \citep{poole2016Exponentiala}, show that expressivity scales exponentially with depth but not width. This means that in the absence of limitations on learning, the best architectures tend to be deeper rather than wider. In cases where learning is restricted, as in greedy layer-wise training, depth and width pose a distinct trade-off: increasing depth increases the number of independently optimized stages and thus makes it more difficult to retain relevant information across-layers, while increasing width adds within-layer capacity. We thus hypothesized that increasing width (instead of depth) could help local learning to compete with end-to-end training. This hypothesis is also motivated by biological learning: compared to modern deep learning algorithms, the brain's networks are much shallower \citep{suzuki2023How, harris2019Hierarchical, dsouza2022Hierarchical} and wider \citep{babadi2014Sparseness, litwin-kumar2017Optimal}. We test our hypothesis in the context of self-supervised learning due to its role as a central framework in large-scale representation pretraining in vision \citep{jing2019Selfsupervised, balestriero2023Cookbook}. 

Specifically, we trained four and eight layer convolutional neural networks (Conv4 and Conv8, respectively) of varying widths on the CIFAR-10 categorization task with the Barlow Twins loss \citep{zbontar2021Barlowa} and then used a k-nearest neighbors ($k$NN) probe to determine how well the learned representations could perform a classification task. (We also performed these experiments with the SimCLR loss \citep{chen2020Simplea} and reached similar conclusions). We trained identical architectures using either greedy layer-wise training or end-to-end backpropagation, enabling a direct comparison between these learning paradigms. Increasing width disproportionately improved representations learned through greedy layer-wise training and progressively closed the performance gap with end-to-end training: in the widest Conv4 network we studied, greedy layer-wise training led to even higher classification performance than did end-to-end training. The deeper Conv8 networks see a similar convergence, although we were unable to train models sufficiently wide to see greedy layer-wise training outperform end-to-end backpropagation. To understand why greedy layer-wise training can compete with end-to-end backpropagation, we analyzed the representational geometry of the trained networks and found that sufficiently wide greedy layer-wise trained networks have increased signal--signal factorization and signal--noise factorization compared to end-to-end trained ones: recent theoretical work \citep{wakhloo2026Neurala} shows that these are the two key properties underlying classifier performance. These results demonstrate that the cost in performance of restricted credit assignment is dependent on network architecture and identify changes in representational geometry as a potential cause. Moreover, this provides a possible explanation for the wide and shallow architectures of our brains: these architectures enable the best performance with localized learning rules, avoiding the need for end-to-end backpropagation of which the neural foundations are currently unknown \citep{lillicrap2020Backpropagation, sacramento2018Dendritic, ellenberger2025Backpropagation}.

\section{Related Work}

Local learning rules are alternatives to backpropagation that restrict credit assignment to a subset of the network. In self-supervised learning, previous work has demonstrated that local learning can create useful representations. For example, GreedyInfoMax uses gradient-isolated modules \citep{lowe2020Putting} and LoCo uses overlapping blocks \citep{xiong2020LoCo} to achieve competitive performance to end-to-end backpropagation. Self-supervised experiments with greedy block-wise training with a Barlow Twins loss also found only slightly worse performance than backpropagation \citep{siddiqui2024Blockwise}. These experiments maintain credit assignment across multiple layers. Similarly, in the realm of supervised learning, \citet{belilovsky2019Greedy} found similar performance to backpropagation in eleven layer networks using greedy training, but required multi-layer auxiliary classifiers. These studies, however, did not systematically study how the choice of model architecture influences the efficacy of local learning.

Other research has studied factors that might inhibit local learning. For example, \citet{bartunov2018Assessing} found that learning rules fall behind as network depth increases. Indeed, a major limitation of local learning is the tendency for local updates to discard information that is useful for later layers \citep{wang2021Revisiting}. We hypothesized that width could provide a solution as increasing width increases the capacity of each layer. However, the effect of width on the performance of these locally trained networks has not been well explored. Here, we investigate whether increasing width may improve greedy layer-wise training more than end-to-end training, allowing it to approach or exceed end-to-end training in performance. 
 
\section{Method}

\subsection{Experimental Design}

We compare end-to-end and greedy layer-wise training with four- and eight-layer convolutional neural networks (Conv4 and Conv8, respectively) using the Barlow Twins loss function \citep{zbontar2021Barlowa} on CIFAR-10 \citep{krizhevsky2009Learning}. Shallow encoders have precedent in SSL: \citet{patacchiola2020SelfSupervised} use Conv4 architectures for the CIFAR-10 and CIFAR-100 datasets with a variety of SSL loss functions while \citet{wang2022Contrastive} use Conv4 with Barlow Twins loss on similar image sets. Conv4 has base ($1\times$) channel count of $32 \rightarrow 64 \rightarrow 128 \rightarrow 256$. The Conv8 maintains the same overall width progression: $32 \rightarrow 32 \rightarrow 64 \rightarrow 64 \rightarrow 128 \rightarrow 128 \rightarrow 256 \rightarrow 256$. This follows the conventional doubling pattern used by Conv4 in \citet{patacchiola2020SelfSupervised} and by standard residual networks \citep{he2016Deep} (though our network does not have residual connections). For our width sweep, we multiply all channel widths by $w\in\{0.25,0.5,1,2,4,8,16\}$ and additionally test $32\times$ for Conv4. Our $0.25\times$ condition matches the channel count of \citet{patacchiola2020SelfSupervised}. See Appendix~\ref{app:architecture} for all architecture specifications.

Models trained end-to-end train for 1,000 epochs, a budget with precedent in Barlow Twins \citep{zbontar2021Barlowa} and CIFAR SSL \citep{ermolov2021Whitening} studies. Greedy-trained Conv4 and Conv8 models allocate 250 and 125 epochs per layer, respectively, matching total number of training epochs. During greedy training of network stage \(l\), all preceding layers are frozen. Augmented views pass through the frozen prefix and active layer; global average pooling then supplies the active layer's projector as in \citet{siddiqui2024Blockwise}. SSL training updated only the parameters of layer \(l\) and its projector. The projector is discarded during representation evaluation. In greedy training, a new projector was initialized after training each layer and then discarded before training the next layer.

\subsection{Hyperparameter Selection}

We selected the optimizer and learning-rate (LR) schedule, projector dimension $q$, and Barlow Twins redundancy coefficient $\lambda$ to maximize validation accuracy of the $k$NN probe applied to end-to-end trained model at representative narrow and wide backbones ($1\times$ and $8\times$). We selected hyperparameters with the explicit goal of maximizing validation performance of the end-to-end-trained models; results for the greedy-trained models did not influence hyperparameter selection. Each hyperparameter screen used a fixed class-balanced 45,000/5,000 split of the CIFAR-10 training set. 

Learning rate was selected through a hyperparameter sweep of literature-supported candidates: Adam at $LR=10^{-3}$ in shallow SSL and small-projector experiments \citep{patacchiola2020SelfSupervised, ghosh2024Harnessing}, Adam at $LR=3\times10^{-3}$ for CIFAR SSL \citep{ermolov2021Whitening}, and LARS with warmup and cosine decay in the original Barlow Twins paper \citep{zbontar2021Barlowa}. We also tested cosine decay alternatives for Adam LR candidates. We evaluated end-to-end training for 200 epochs during the hyperparameter search. Adam with constant $LR=10^{-3}$ achieved the highest validation accuracy for end-to-end-trained models at both widths and was used for the following projector sweep. Full details are in Appendix~\ref{app:hyperparam}. An additional 1,000 epoch end-to-end sweep with Adam $LR=10^{-3}$ with cosine decay improved performance at larger widths, although greedy layer-wise training $k$NN accuracy remained higher. 

\citet{ghosh2024Harnessing} show that small projectors can be effective when the redundancy penalty is adjusted, motivating joint selection of $q$ and $\lambda$. We chose to compare two-layer heads $h\rightarrow q\rightarrow q$ due to precedent from previous Conv4 Barlow Twins \citep{wang2022Contrastive} and CIFAR Barlow Twins \citep{chapman2024Unconstrained} implementations. We use heads $h\rightarrow q\rightarrow q$, where $h$ is the encoder feature dimension, with equal hidden and output dimensions as in \citet{chapman2024Unconstrained}. We tested $q\in\{256,1024,4096\}$, a geometrically spaced subset of the dimensions evaluated by \citet{ghosh2024Harnessing}. For each $q$, we tested three coefficients: $0.0051$ from the original Barlow Twins implementation \citep{zbontar2021Barlowa}; the corresponding CIFAR-10 configuration from \citet{ghosh2024Harnessing} ($0.02$, $0.002$, and $0.0004$, respectively); and $0.0051(8192/q)$, an inverse-dimension candidate motivated by \citet{ghosh2024Harnessing}'s recommended scaling. These nine configurations were each trained end-to-end for 500 epochs at both widths. We ranked configurations by terminal validation $k$NN accuracy averaged across widths, then evaluated the two leading configurations by training new networks randomly initialized with a different RNG seed. The selected setting was $q=256$ and $\lambda=0.1632$ (Tables~\ref{tab:bt_projector_lambda} and~\ref{tab:bt_projector_confirm}).

With the selected projector and coefficient, a second 200-epoch end-to-end screen compared constant-LR Adam at $10^{-3}$ and $3\times10^{-3}$, cosine-decayed Adam at $10^{-3}$, and LARS with warmup and cosine decay. Constant-LR Adam at $10^{-3}$ had the highest mean validation accuracy (Table~\ref{tab:bt_lr}). 

Hyperparameters for model training with the SimCLR loss were determined by a similar hyperparameter validation approach, which is detailed in Appendix~\ref{app:simclr}.

\begin{figure}[h]
\begin{center}
\setlength{\fboxsep}{4pt}
\setlength{\fboxrule}{0.5pt}
\includegraphics[width=0.8\textwidth]{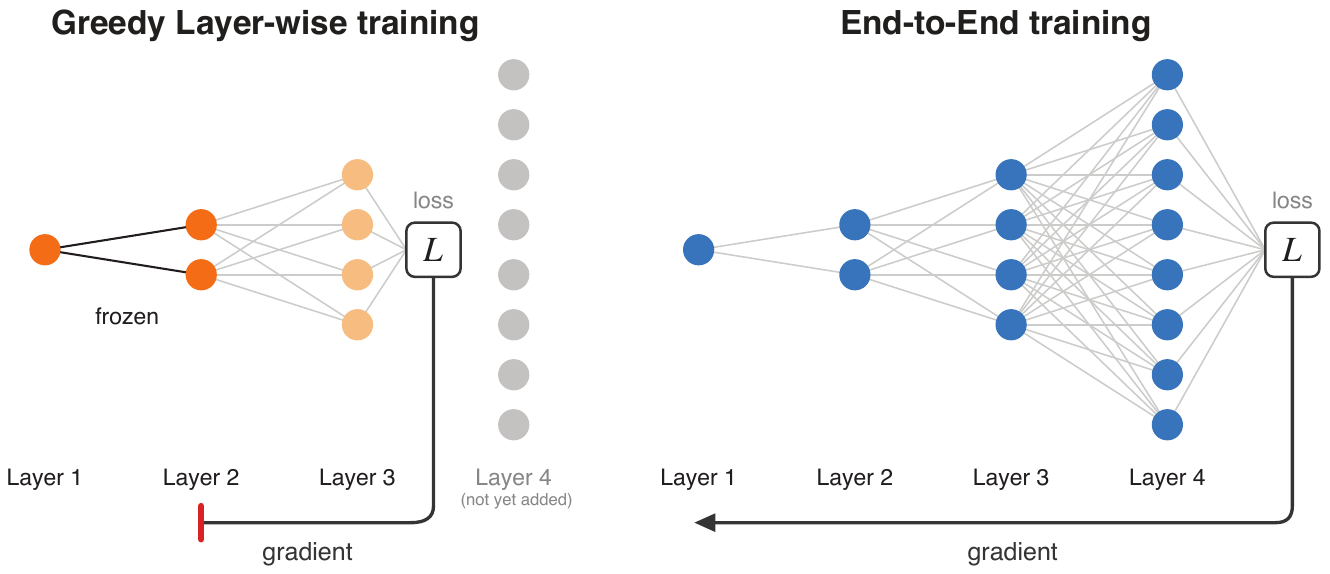}
\end{center}
\caption{\textbf{Greedy layer-wise vs.\ end-to-end training.}
\textit{Left:} Demonstration of greedy training. As an example, we show what happens while training Layer 3 (the same procedure is applied to the other layers). In this example, Layers 1--2 are frozen, only Layer 3 is trained, and the gradient stops at Layer 2. Layer 4 is not yet added to the network; it will be trained once Layer 3 is trained.
\textit{Right:} End-to-end training. All layers are trained jointly, and the gradient reaches every layer.}\end{figure}

\subsection{Representational Geometry}
\label{method:rep geo}

We adapted the representational geometry framework of \citet{wakhloo2026Neurala} to characterize the geometry of encoder representations. We used all 10,000 unaugmented CIFAR-10 test images with class identity as the latent variable. For encoder activations $X$ and class identity matrix $Z$, we calculated the following quantities:

\[
\mathbf{\Psi}=\frac{X^\top X}{N},
\qquad
\mathbf{\Omega}=\frac{Z^\top Z}{N},
\qquad
\mathbf{\Phi}=\frac{X^\top Z}{N},
\]
where $\mathbf{\Psi}$ and $\mathbf{\Omega}$ are the representation
and latent variable covariance matrices, and $\mathbf{\Phi}$ is their
cross-covariance. Signal--signal factorization $f$ (SSF) and signal--noise
factorization $s$ (SNF) were:
\begin{equation}
f=
\frac{
\left[\operatorname{Tr}
(\mathbf{\Phi}\mathbf{\Phi}^{\top})\right]^2
}{
\operatorname{Tr}(\mathbf{\Omega})\,
\operatorname{Tr}\!\left(
\mathbf{\Phi}^{\top}\mathbf{\Phi}
\mathbf{\Omega}^{-1}
\mathbf{\Phi}^{\top}\mathbf{\Phi}
\right)
}.
\end{equation}

\begin{equation}
s=
\frac{
\left[\operatorname{Tr}
(\mathbf{\Phi}\mathbf{\Phi}^{\top})\right]^2
}{
\operatorname{Tr}(\mathbf{\Omega})\,
\operatorname{Tr}\!\left[
\mathbf{\Phi}^{\top}
\left(
\mathbf{\Psi}-
\mathbf{\Phi}\mathbf{\Omega}^{-1}\mathbf{\Phi}^{\top}
\right)
\mathbf{\Phi}
\right]
}.
\end{equation}

SSF is a measure of how evenly differences between class means are distributed across independent directions in the representation. Higher SSF indicates that class-related variation is spread more evenly, rather than concentrated in a few dominant directions. SNF measures the strength of these differences relative to within-class variation along the same directions. Higher SNF indicates that within-class variation is smaller relative to class signal. Theoretical work from \citet{wakhloo2026Neurala} shows that these two quantities are important for determining how well classifiers can generalize to new samples.

\section{Results}

\subsection{Increasing width narrows the greedy training performance gap}

\begin{figure}[t]
    \centering
    \includegraphics[width=\textwidth]{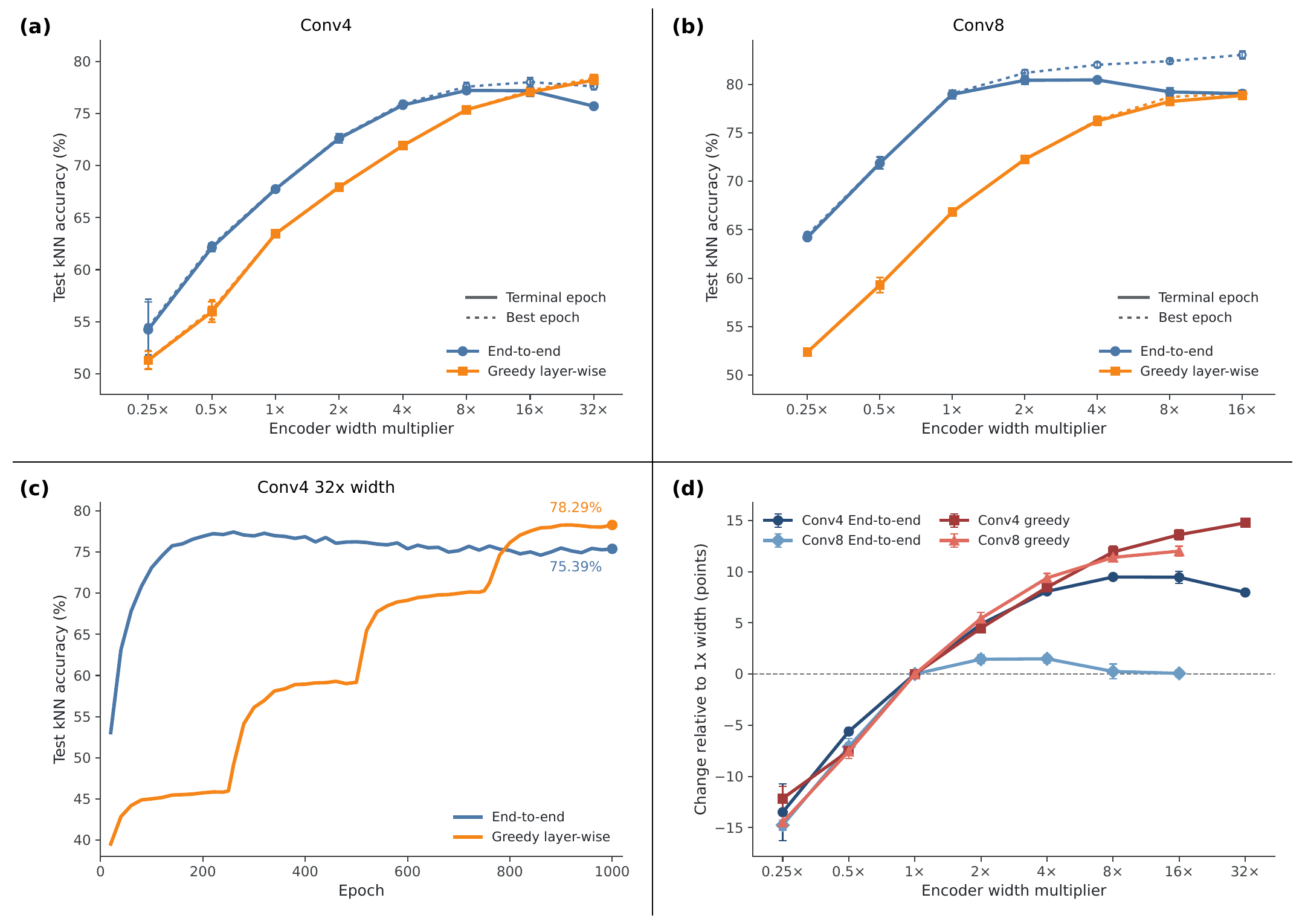}
    \caption{
    \textbf{Increased network width narrows the performance gap between greedy layer-wise and end-to-end training.}
    Conv4 and Conv8 networks trained with Barlow Twins loss on the CIFAR-10 dataset.
    \textbf{(a,b)} Terminal and best-epoch test $k$NN accuracy across widths for Conv4 (a) and Conv8 (b). Solid lines indicate terminal accuracy; dotted lines indicate the highest recorded accuracy.
    \textbf{(c)} Example test $k$NN accuracy training trajectory for Conv4 at $32\times$ width.
    \textbf{(d)} Change in terminal accuracy relative to each method's $1\times$ baseline.}
    \label{fig:width_scaling}
\end{figure}

We first asked if increasing width changes the relative performance of identical networks with end-to-end versus greedy layer-wise training. For 4 layer CNNs (Conv4) trained to minimize the Barlow Twins loss and then evaluated on categorization performance with a $k$NN probe, increasing width improved the performance of both end-to-end and greedy layer-wise trained models. Improvement was larger for greedy layer-wise training (Fig.~\ref{fig:width_scaling}a, b). At smaller widths, end-to-end training achieved higher accuracy than greedy layer-wise training, and the performance gap decreased as width increased. At $16\times$ width, end-to-end training and greedy layer-wise training were almost identical, and at $32\times$ width, greedy layer-wise training achieved higher accuracy than end-to-end training. The reversal -- wherein end-to-end training led to higher accuracy for narrower models but greedy layer-wise training led to higher accuracy for $32 \times$ width models -- was present in all three training runs we performed, each of which had a different RNG seed.

After noting that categorization performance could decline in the later stages of training, we sought to compare the highest recorded test $k$NN accuracy achieved by each model. We then compared those ``best epochs'' across models and learning paradigms. This comparison demonstrates that the improved performance achieved by greedy training at $32\times$ width is partially due to a deterioration in performance of the model trained end-to-end. At $32\times$ width, mean best-epoch accuracy for end-to-end training was only slightly below that of greedy training. The $32\times$ width example learning trajectory demonstrates this: accuracy of the model trained end-to-end peaked at $77.43\%$ at epoch 260 and ended at $75.39\%$, whereas accuracy of the model with greedy training achieved its maximum accuracy of $78.29\%$ at epoch 920 and remained around that value (Fig.~\ref{fig:width_scaling}c).

For the Conv8 architecture, we found a similar result. At narrow widths, greedy layer-wise training achieves much lower performance than end-to-end training whereas at $16\times$ width, greedy layer-wise training achieves a categorization accuracy that is only slightly below that of end-to-end training. In the widest Conv8 networks, the maximum accuracy achieved by the networks during training remained higher in end-to-end trained networks even though their terminal epoch accuracy converged with that of the greedy trained networks (Fig.~\ref{fig:width_scaling}b). We therefore find that the convergence in performance for Conv8 is representative of both an uneven improvement in greedy layer-wise training and a degradation of performance in the later stages of end-to-end training. Nonetheless, we see a substantial and disproportionate improvement with network widening in greedy layer-wise training as compared to end-to-end training. Specifically, widening Conv4 from $1\times$ to $32\times$ increased terminal accuracy by 14.76 points for greedy training versus 7.97 points for end-to-end training. Even more dramatically, widening Conv8 from $1\times$ to $16\times$ produced a gain of 12.01 points in the greedy-trained networks compared to just 0.07 points for the end-to-end-trained networks (Fig.~\ref{fig:width_scaling}d).

\subsection{Barlow Twins loss is an imperfect proxy for downstream classification accuracy}

Given that end-to-end training should generally be better at minimizing the training loss than greedy layer-wise training, we were curious about why greedy layer-wise training was able to show comparable performance in wide networks. We hypothesized that this effect could arise due to the fact that minimizing training loss (Barlow Twins loss) is not quite the same thing as maximizing categorization performance with the $k$NN probe.

\begin{figure}[t]
    \centering
    \includegraphics[width=\textwidth]{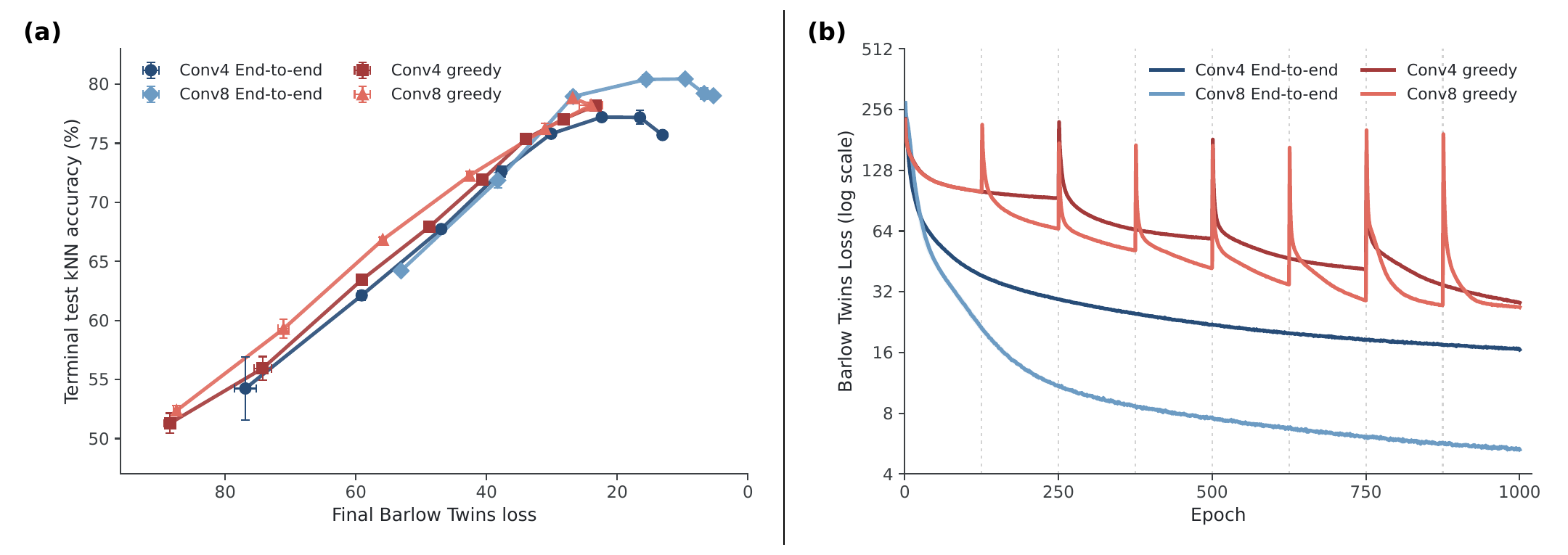}
    \caption{
    \textbf{Lower Barlow Twins loss does not necessarily imply higher categorization accuracy.}
    \textbf{(a)} Terminal test $k$NN accuracy versus final training loss across network widths. Each point represents one width; connecting lines follow increasing width. The loss axis decreases from left to right.
    \textbf{(b)} Training loss curves for models with $16\times$ width for both Conv4 and Conv8 architectures with either greedy layer-wise training or end-to-end (E2E) training. The loss axis (y) is logarithmic. Dotted vertical lines indicate layer additions for greedy training.
    }
    \label{fig:loss_accuracy}
\end{figure}

To test this hypothesis, we investigate the relationship between the Barlow Twins loss and the categorization accuracy over training. Across all model widths, lower Barlow Twins loss was broadly associated with a higher $k$NN accuracy. This trend of wider models having lower losses and higher $k$NN accuracies was observed for both depths and training methods (Fig.~\ref{fig:loss_accuracy}a). However, at the lowest values of the Barlow Twins loss, this trend ceased to hold. In Conv4 at $32\times$ width, the mean final loss was higher for greedy training than end-to-end training, despite the higher terminal accuracy of the greedy-trained model. Similarly, Conv8 at $16\times$ width reached nearly matching terminal accuracies for the two training methods despite substantial differences in their final loss values (Fig.~\ref{fig:loss_accuracy}a).

Widening the models largely improved their accuracy. Beyond a certain point, however, further increases in width sometimes reduced terminal loss while accuracy saturated or declined: this effect was especially prevalent in models with end-to-end training (Fig.~\ref{fig:loss_accuracy}a). Overall, we find that lower Barlow Twins loss does not consistently indicate better categorization performance.

\subsection{Wide networks with greedy layer-wise training exhibit distinct class-related representational geometry}

To understand the mismatch between Barlow Twins loss and categorization accuracy, we looked to representational geometry to determine how differently-trained networks organize class-related information. We measured signal--signal factorization (SSF) and signal--noise factorization (SNF) across network widths and training methods in Conv4. SSF measures how evenly differences between classes are spread out in representation space while SNF describes how strong these class differences are relative to variation within-class along those same directions. Recent theoretical work indicates that these are two key quantities that determine how well network representations can enable categorization tasks \citep{wakhloo2026Neurala}. See Method~\ref{method:rep geo} for more details.

\begin{figure}[t]
    \centering
    \includegraphics[width=\textwidth]{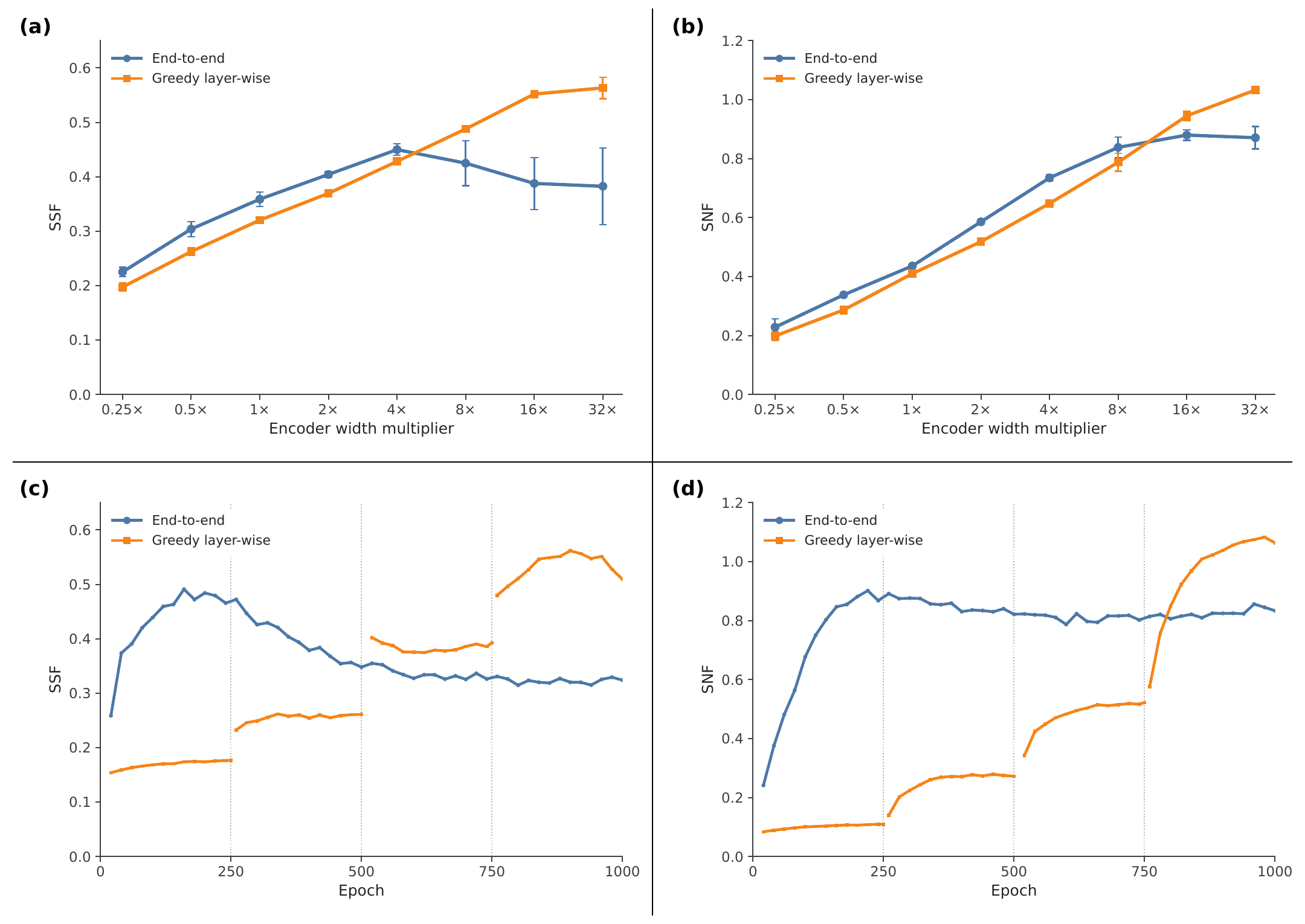}
    \caption{
    \textbf{Representational geometry across network widths and training methods.}
    Signal--signal factorization (SSF) and signal--noise factorization (SNF) for Conv4 networks trained to minimize Barlow Twins loss on CIFAR-10.
    \textbf{(a,b)} Final encoder-layer SSF and SNF across widths.
    \textbf{(c,d)} SSF and SNF over 1,000 epochs for Conv4 $32\times$ width.
    }
    \label{fig:representation_geometry}
\end{figure}

This analysis revealed different trends in SSF and SNF for models with end-to-end vs greedy layer-wise training. For models trained end-to-end, SSF first increased with increasing width, before decreasing as width exceeded $4\times$. In contrast, SSF in greedy-trained networks continued to increase across widths, exceeding the SSF of models trained end-to-end for widths $>8\times$ (Fig.~\ref{fig:representation_geometry}a).

SNF generally increased with width for models trained with either method, but showed different trends at the highest widths. For models trained end-to-end, SNF increased up to $16 \times$ width before saturating. In contrast, SNF for greedy-trained models continued to increase across the tested widths. Compared to greedy layer-wise training, end-to-end-trained networks had higher SNF through $8\times$ width. However, greedy-trained networks had higher SNF at $16\times$ and $32\times$ width (Fig.~\ref{fig:representation_geometry}b). 

Thus, the widest greedy-trained networks had higher SSF and SNF to accompany their improved categorization accuracy. Notably, the width values for which SSF or SNF of the greedy-trained networks exceed those of the end-to-end-trained ones do not exactly match those at which the $k$NN accuracies cross over. For example, SSF for greedy layer-wise trained models is already higher than for end-to-end trained models at $8\times$ width, whereas the $k$NN accuracy of the greedy-trained network is lower for this width. This implies that improved SSF alone is not sufficient to explain the increased categorization accuracy of the widest greedy-trained networks.

We next asked if the geometric advantage in wide greedy-trained networks was due to improvements in greedy training or a deterioration in end-to-end training. We evaluated changes in representational geometry over the course of training in the $32\times$ width networks. For models trained end-to-end, SSF reached its maximum at epoch 160 and continuously decreased until epoch 1,000. In contrast, SSF for models with greedy layer-wise training increased as each additional layer was trained, eventually surpassing the maximum achieved by end-to-end training (Fig.~\ref{fig:representation_geometry}c). SNF followed a similar trend: end-to-end training achieved its maximum at epoch 220 while for greedy layer-wise training, SNF increased with each added layer, eventually surpassing the SNF achieved by end-to-end training (Fig.~\ref{fig:representation_geometry}d). SSF and SNF for greedy layer-wise training at $32\times$ width surpass both the maximum and the endpoint (final training epoch) SSF and SNF values achieved by end-to-end training. 

These results establish that greedy-trained networks may organize class-related information more favorably than end-to-end trained networks, identifying class-related geometry as a potential explanation for the difference in representational quality between end-to-end and greedy training.  

\begin{figure}[t]
    \centering
    \includegraphics[width=\textwidth]{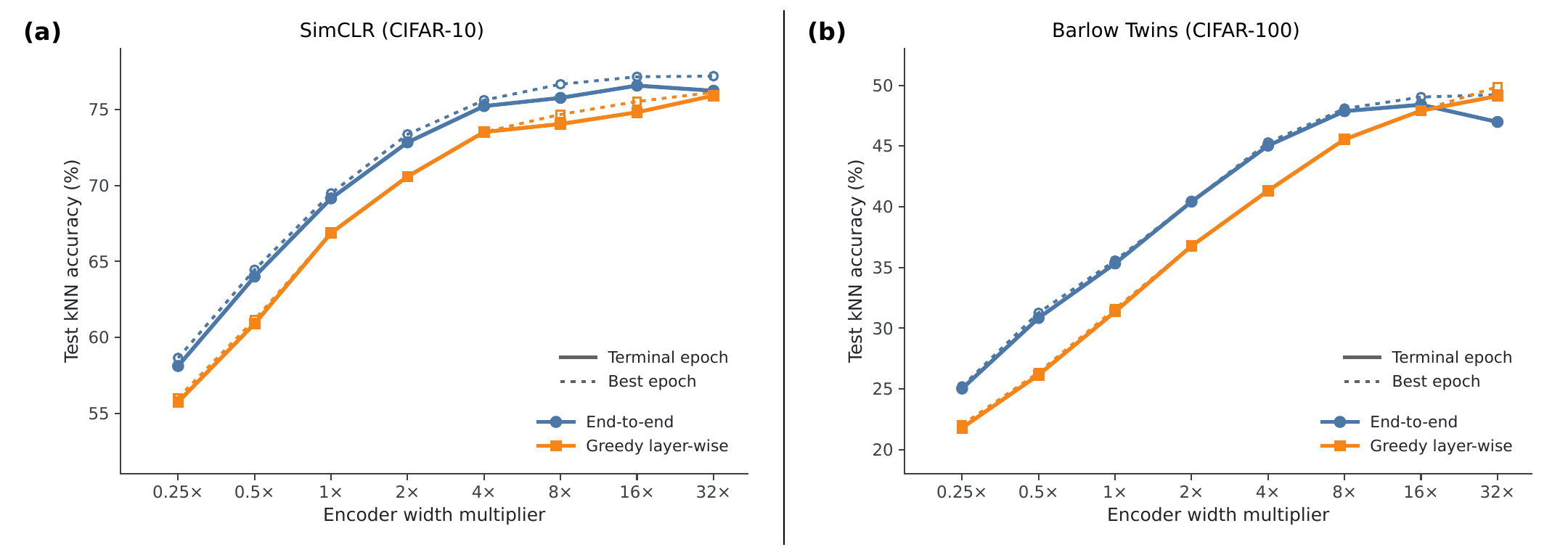}
    \caption{
    \textbf{Width-dependent performance trends extend to another self-supervised loss function and dataset.}
    Terminal and best-epoch test $k$NN accuracy for Conv4 networks trained end-to-end or greedily layer-wise on a single seed.
    \textbf{(a)} $k$NN categorization accuracy for Conv4 models trained to minimize SimCLR loss on CIFAR-10 across widths from $0.25\times$ to $32\times$. \textbf{(b)} $k$NN categorization accuracy for Conv4 models trained to minimize Barlow Twins loss on the CIFAR-100 dataset across widths from $0.25\times$ to $32\times$. Solid lines indicate terminal accuracy; dotted lines indicate the highest recorded accuracy at full encoder depth. 
    }
    \label{fig:generalization}
\end{figure}

\subsection{Width effects generalize across loss functions and datasets}

Finally, we tested if our results extend beyond models trained to minimize Barlow Twins loss on the CIFAR-10 dataset. To achieve that goal, we trained models to minimize SimCLR loss \citep{chen2020Simplea} on the CIFAR-10 dataset and trained our Barlow Twins models on the CIFAR-100 dataset.

Models trained with SimCLR loss on CIFAR-10 behaved similarly to those trained with Barlow Twins loss: increasing width provided a larger performance gain for models with greedy layer-wise training as compared to end-to-end training, with the categorization accuracy gap shrinking as width increased. Different from our Barlow Twins results, we did not see a complete convergence: even for the largest width we studied, end-to-end training retained a small performance advantage (Fig.~\ref{fig:generalization}a).

We next trained Conv4 models to minimize Barlow Twins loss on the CIFAR-100 dataset. Similar to our findings with CIFAR-10, increasing width caused a greater performance gain for greedy-trained models than for end-to-end-trained ones, and greedy training led to higher accuracy for the widest networks we studied (Fig.~\ref{fig:generalization}b).

\section{Discussion}

We investigated the effect of width on the performance of networks trained with greedy layer-wise training, asking for what architectures a network trained with restricted credit assignment can produce representations comparable in performance to networks trained end-to-end. Across four and eight layer convolutional neural networks trained with the Barlow Twins loss on the CIFAR-10 dataset, increasing network width disproportionately improved the categorization accuracy (assessed with a $k$NN probe) of greedy layer-wise trained networks compared to end-to-end trained networks. Greedy training matched or even surpassed end-to-end training in terms of model performance for the widest models. Additional experiments showed a similar performance convergence for networks trained with SimCLR contrastive loss, and with a different dataset (CIFAR-100 instead of CIFAR-10), thus showing that our findings are not specific to Barlow Twins loss or to the CIFAR-10 dataset.

Our representational geometry analysis provides a possible explanation for these differences. Greedy layer-wise trained networks reached higher signal--signal factorization and signal--noise factorization in our widest networks, indicating that these representations distributed class related variation more evenly and had stronger class signal relative to within-class variation. This is consistent with the theoretical work of \citet{wakhloo2026Neurala}, who show that lower classification error is possible as SSF and SNF increase,  as well as related work showing that categorization error is related to the overlap between signal and noise directions in representation space \citep{sorscher2022Neural}.

Why greedy layer-wise trained networks produce representations with better SSF and SNF than ones with end-to-end training is not fully known. However, one possibility is that freezing earlier layers may preserve class-related structure that end-to-end optimization might alter over lengthy training on SSL loss functions (which do not explicitly reinforce categorical representations). This hypothesis is consistent with our observation of SSF declining over the latest epochs during end-to-end training. Future research should further explore how local learning, network width, and layer freezing shape representational geometry.

Our results suggest that increasing network width can compensate somewhat for a lack of global credit assignment. This finding is particularly relevant for the exploration of biologically plausible alternatives to backpropagation. Backpropagation is often cited as biologically implausible due to its globally coordinated credit assignment. Exploring relatively shallow, wide networks with local learning is a biologically motivated alternative: the brain has relatively few processing stages (shallow architecture) \citep{suzuki2023How, harris2019Hierarchical, dsouza2022Hierarchical} with massive neural populations (wide architecture) \citep{babadi2014Sparseness, litwin-kumar2017Optimal}. Thus, our results motivate the further exploration of wide, shallow networks optimized with local learning rules. More broadly, our results support the investigation of width as a potential factor to influence the performance of biologically plausible learning rules.

Our work has several limitations. First, our tested architectures are limited to relatively shallow convolutional networks. Greater depth or architectural differences, such as residual connections, could result in a different relationship between width and performance in greedy layer-wise training. At the same time, our main finding generalized well between architectures of quite different depth (4 vs 8 layers): in both cases, increasing width shrinks the gap between networks trained end-to-end with backprop vs greedy layer-wise methods. We thus anticipate that this same effect would apply more broadly. We hope that future work explores these ideas in a wider range of contexts, broadening our perspective on when and why localized credit assignment may produce representations of comparable quality to end-to-end backpropagation of error. 

Next, our experiments were limited to two loss functions and two datasets. We primarily studied networks trained with the Barlow Twins loss function on the CIFAR-10 dataset. At the same time, our primary results also held for a different loss function and dataset, suggesting that our findings are likely generalizable across additional datasets and choices of loss function. Finally, our study is limited to self-supervised learning, and we did not investigate whether a similar width-dependent result is present in other contexts such as supervised learning. Nevertheless, the central role of self-supervised learning in representation pretraining makes it an important setting in which to study alternative training paradigms.

Overall, we identified an architectural condition where greedy layer-wise training can produce representations of comparable quality to those learned through end-to-end backpropagation. We show that increased width benefits greedy layer-wise training, narrowing the performance gap with end-to-end backpropagation. We point to differences in representational geometry as a potential reason for improved greedy layer-wise representations. Together, these results motivate width as a primary variable determining the efficacy of local learning paradigms.

\subsection*{AI use statement}

We used Generative AI tools to assist in summarizing and analyzing existing literature. In addition, Generative AI tools were used to assist in code implementation, debugging, and refactoring as well as to polish writing. Generative AI tools were not used in research ideation and the rest of the required disclosure tasks are not applicable to this work. We have reviewed all AI-assisted work by manually reviewing all literature. We also reviewed all LLM-generated code and verified and tested it for correctness. We take responsibility for the final content of this work, including text, claims or artifacts produced with the aid of generative AI.

\subsection*{Ethics statement}

This work does not involve human subjects, potentially harmful insights, user studies, dataset releases, or applications in sensitive domains. We do not identify any concerns related to bias, fairness, privacy, or security arising from this research. No conflicts of interest or external sponsorship relevant to this work exist. We believe this submission raises no significant ethical concerns beyond those inherent to standard machine learning research.

\subsection*{Reproducibility statement}

All experimental design details are described in the Methods and Appendix. Appendix~\ref{app:architecture} specifies the architectures, training procedures, data transformations, and evaluation protocols. Appendices~\ref{app:barlow} and~\ref{app:simclr} detail the objectives and hyperparameter selection, and Appendix~\ref{app:representational_geometry} describes the geometry calculations. All code used to produce the reported results will be linked in a GitHub repository after the double blind review period. Additional reproducibility details are included in Appendix~\ref{app:reproducibility}.

\subsubsection*{Author Contributions}

SM: Research ideation, investigation, analysis, writing.

JZ: Research ideation, analysis, mentoring, funding, resources, editing and reviewing.

\subsubsection*{Acknowledgments}

This work was supported by Grant $\#$ R01-EY036594 from the National Institutes of Health, an Unrestricted grant from Research to Prevent Blindness to the Jules Stein Eye Institute, Grant $\#$ FA9550-26-1-B113 from the Air Force Office of Scientific Research, a CIFAR Fellow Award for Learning in Machines and Brains, and the Joan and Jerome Snyder Chair in Vision Science at UCLA.

\newpage

\bibliography{0911_bib}
\bibliographystyle{iclr2027_conference}

\newpage

\appendix
\section{Appendix}

\subsection{Barlow Twins}
\label{app:barlow}

\subsubsection{Loss Function}
\label{app:barlow_loss}
The Barlow Twins loss function compares the projected representations of two augmented views of the same underlying image \citep{zbontar2021Barlowa}. For an encoder output with $h$ channels, the projector is a two layer MLP of shape $h\rightarrow q\rightarrow q$, with bias-free linear layers and BatchNorm followed by ReLU after the first linear layer only. The final layer has a linear activation function and does not use BatchNorm. We use $q=256$, selected from $\{256,1024,4096\}$ through a joint screen of projector dimension and redundancy coefficient based on mean end-to-end validation accuracy at $1\times$ and $8\times$ width (Appendix~\ref{app:hyperparam}).

The loss function is:
\begin{equation}
\mathcal L_{\mathrm{BT}}=
\underbrace{\sum_{i=1}^{q}(1-C_{ii})^2}_{\mathcal L_{\mathrm{inv}}}
+\lambda\underbrace{\sum_{i\ne j}C_{ij}^2}_{\mathcal L_{\mathrm{red}}},
\qquad \lambda=0.1632.
\label{eq:app_bt_loss}
\end{equation}

Here, $C\in\mathbb R^{q\times q}$ is the cross-correlation matrix, computed across $B$ images, between the projected representations of the two views. Minimizing the first term in the loss encourages corresponding projected features in the two views to have a correlation approaching 1, thereby encouraging the representation to be invariant to image augmentation. In contrast, minimizing the second term of the loss discourages the representations of different features from encoding redundant information. The coefficient $\lambda$ controls the relative weight of the redundancy-reduction term.  We use $\lambda=0.1632$, selected jointly with the projector dimension to maximize mean end-to-end validation accuracy at $1\times$ and $8\times$ width (Appendix~\ref{app:hyperparam}).

\subsubsection{Encoder architecture}
\label{app:architecture}

Each encoder layer contains a bias-free $3\times3$ convolution with padding $1$, followed by BatchNorm and ReLU. There are no residual connections. Conv8 has an additional stride-one layer at each channel scale while preserving Conv4's downsampling. Global average pooling (GAP) produces one feature per channel for the projection head and for evaluation. 

\subsubsection{Image Augmentation Pipeline}

Each of the two independently augmented views underwent random resized cropping to $32\times32$. Cropping area size was randomly chosen with a minimum of 20\%. Resizing used bicubic interpolation with antialiasing. We then applied horizontal flip with probability 0.5, color jitter with probability 0.8, and grayscale conversion with probability 0.2. Jitter magnitudes were 0.4 for brightness, 0.4 for contrast, 0.2 for saturation, and 0.1 for hue. These settings follow the original Barlow Twins implementation \citep{zbontar2021Barlowa}.

Consistent with other shallow-network SSL implementations, we omitted Gaussian blur and solarization \citep{patacchiola2020SelfSupervised}. We increased minimum crop area from 8\% in the original paper to 20\%, a setting also used in CIFAR-10 implementations of \citet{khosla2020supervised} and \citet{ermolov2021Whitening}. This seemed appropriate given the small size of the CIFAR-10 and CIFAR-100 images ($32\times32$).

\subsubsection{Evaluation}
\label{app:evaluation}

Evaluation uses features extracted from global average pooling (GAP) of the encoder's convolutional layers. The projector was excluded from evaluation. For our $k$NN evaluation, we build a memory bank of feature vectors and labels for all 50,000 training images. We then, for each test image, selected the 200 memory-bank feature vectors with the largest cosine similarity to the test image's feature vector. Unweighted majority vote over the category labels associated with those 200 feature vectors was used to classify the test image. Reported accuracy is top-1 classification accuracy over all 10,000 test images. Evaluation occurred every 20 epochs and at the end of each stage of greedy layerwise training. Terminal accuracy was calculated at epoch 1,000. Best-epoch accuracy was the maximum accuracy obtained over all epochs.

\subsubsection{Hyperparameter Selection}
\label{app:hyperparam}

To select the final optimizer, learning rate, and projector dimensions, we used terminal end-to-end Conv4 validation accuracy at $1\times$ and $8\times$ width, averaged. We optimized all hyperparameter choices for end-to-end performance and then used identical hyperparameters in our greedy layer-wise networks so as to conservatively estimate the relative performance of greedy training versus end-to-end backpropagation. Each hyperparameter screen used a class-balanced split of the CIFAR-10 training set: 45,000 images for SSL training and the $k$NN memory bank, and 5,000 separate images for validation. Batch size was 512.

We performed an initial screen for each candidate optimizer and learning rate combination for 200 epochs with seed 0, a projector $h\rightarrow4096\rightarrow256$, and $\lambda=0.0051$. After selecting the projector dimension and $\lambda$ in a separate screen, we repeated a learning rate screen to ensure that these changes did not alter our learning rate choice (Table~\ref{tab:bt_lr}).

All Adam candidates used zero weight decay. Cosine-decayed Adam reached zero learning rate at the end of training. LARS used peak rate $0.2(B/256)=0.4$ for matrix weights and $0.0048(B/256)=0.0096$ for biases and BatchNorm parameters, momentum 0.9, trust coefficient $10^{-3}$, and weight decay $10^{-6}$ as was done in the original Barlow Twins paper. The schedule used two warmup epochs followed by cosine decay toward $10^{-3}$ times the peak rate. Constant Adam at $10^{-3}$ ranked first at both widths Table~\ref{tab:bt_initial_lr}.

Noting that performance of networks trained end-to-end degraded at larger widths, we performed an additional width sweep with Adam $LR=10^{-3}$ with cosine decay to see if keeping a constant learning rate contributed to this degradation. Although end-to-end training improved its terminal performance at $32\times$ width, greedy layer-wise training remained higher (Table~\ref{tab:bt_lr_schedule}).

\begin{table}[htbp]
\centering\small
\caption{Initial end-to-end optimizer screen with terminal validation $k$NN accuracy. Reported values are top-1 classification accuracy on CIFAR-10 using a $k$NN probe.}
\label{tab:bt_initial_lr}
\begin{tabular}{@{}lrrr@{}}
\toprule
Recipe & $1\times$ & $8\times$ & Mean\\
\midrule
Adam $10^{-3}$, constant & \textbf{67.12} & \textbf{75.36} & \textbf{71.24}\\
Adam $3\times10^{-3}$, cosine & 66.52 & 73.88 & 70.20\\
Adam $10^{-3}$, cosine & 65.82 & 73.82 & 69.82\\
Adam $3\times10^{-4}$, cosine & 62.20 & 71.74 & 66.97\\
Adam $10^{-4}$, cosine & 57.78 & 68.88 & 63.33\\
LARS, warmup + cosine & 58.06 & 66.90 & 62.48\\
\bottomrule
\end{tabular}
\end{table}

\begin{table}[t]
\centering
\small
\caption{Additional learning-rate schedule test for end-to-end learning with Adam $10^{-3}$ cosine and constant. Results of greedy layerwise training are included for reference. Reported values are terminal validation $k$NN probe accuracy on CIFAR-10.}
\label{tab:bt_lr_schedule}
\begin{tabular}{@{}lrrr@{}}
\toprule
Width & Adam $10^{-3}$ const (end-to-end)
      & Adam $10^{-3}$ cosine (end-to-end)
      & Adam $10^{-3}$ const (greedy) \\
\midrule
$0.25\times$ & 55.37 & 55.52 & 50.40 \\
$0.5\times$  & 61.70 & 61.62 & 56.95 \\
$1\times$    & 67.69 & 66.58 & 63.60 \\
$2\times$    & 72.11 & 71.36 & 67.70 \\
$4\times$    & 75.85 & 75.43 & 71.98 \\
$8\times$    & 77.32 & 77.88 & 75.37 \\
$16\times$   & 77.83 & 77.88 & 77.03 \\
$32\times$   & 75.39 & 77.08 & 78.29 \\
\bottomrule
\end{tabular}
\end{table}

To determine projector dimensions and $\lambda$ value, we evaluated $h\rightarrow q\rightarrow q$ projectors with $q\in\{256,1024,4096\}$, a geometrically spaced subset of dimensions studied by \citet{ghosh2024Harnessing}. For each dimension, we tested $\lambda=0.0051$ from the original Barlow Twins implementation; the corresponding public CIFAR-10 FastSSL setting ($0.02,0.002,0.0004$); and $0.0051(8192/q)$, per \citet{ghosh2024Harnessing}'s recommendation. Each of the nine configurations was trained end-to-end for 500 epochs at both widths, using seed 0 and constant Adam at $10^{-3}$. We then evaluated the two configurations with the highest mean terminal validation accuracy (Table~\ref{tab:bt_projector_lambda}).

\begin{table}[htbp]
\centering
\small
\caption{Barlow Twins projector/lambda screen for models trained end-to-end. Reported values are terminal validation $k$NN probe accuracy on CIFAR-10.}
\label{tab:bt_projector_lambda}
\begin{tabular}{rrlrrr}
\toprule
$q$ & $\lambda$ & Candidate source & $1\times$ & $8\times$ & Mean\\
\midrule
256  & 0.0051 & Fixed          & 64.78 & 72.62 & 68.70\\
256  & 0.0200 & Ghosh/FastSSL  & 65.40 & 74.92 & 70.16\\
\textbf{256}  & \textbf{0.1632} & \textbf{Scaled} & \textbf{66.16} & \textbf{75.98} & \textbf{71.07}\\
\midrule
1024 & 0.0051 & Fixed          & 66.48 & 74.64 & 70.56\\
1024 & 0.0020 & Ghosh/FastSSL  & 66.14 & 74.14 & 70.14\\
1024 & 0.0408 & Scaled         & 66.78 & 74.90 & 70.84\\
\midrule
4096 & 0.0051 & Fixed          & 66.88 & 73.68 & 70.28\\
4096 & 0.0004 & Ghosh/FastSSL  & 66.30 & 73.80 & 70.05\\
4096 & 0.0102 & Scaled         & 65.68 & 74.26 & 69.97\\
\bottomrule
\end{tabular}
\end{table}

\begin{table}[htbp]
\centering
\small
\caption{Additional-seed confirmation of the two leading projector/lambda
configurations for end-to-end training. Reported values are terminal validation $k$NN probe accuracy on CIFAR-10.}
\label{tab:bt_projector_confirm}
\begin{tabular}{rrrrrr}
\toprule
Width & $q$ & $\lambda$ & Seed 0 & Seed 1 & Mean $\pm$ SD\\
\midrule
$1\times$ & \textbf{256}  & \textbf{0.1632} & 66.16 & 65.90 & $\mathbf{66.03\pm0.18}$\\
$1\times$ & 1024 & 0.0408 & 66.78 & 65.26 & $66.02\pm1.07$\\
$8\times$ & \textbf{256}  & \textbf{0.1632} & 75.98 & 75.78 & $\mathbf{75.88\pm0.14}$\\
$8\times$ & 1024 & 0.0408 & 74.90 & 75.04 & $74.97\pm0.10$\\
\bottomrule
\end{tabular}
\end{table}

We selected $q=256$ and $\lambda=0.1632$, which had the highest two-width mean in the first screen and kept the higher mean after the additional-seed comparison (Table~\ref{tab:bt_projector_confirm}). To check if this result interacted with our learning rate choice, we repeated a 200-epoch end-to-end optimizer screen at both widths using $q=256$ and $\lambda=0.1632$ with seed 0 using a variety of learning rates and schedulers. Constant Adam at $10^{-3}$ had the highest mean terminal validation accuracy (Table~\ref{tab:bt_lr}).

\begin{table}[htbp]
\centering
\small
\caption{Barlow Twins optimizer/schedule screen. Entries are terminal
validation $k$NN probe accuracy (\%) on CIFAR-10 after end-to-end training.}
\label{tab:bt_lr}
\begin{tabular}{llrrr}
\toprule
Optimizer & Schedule & $1\times$ & $8\times$ & Mean\\
\midrule
\textbf{Adam $10^{-3}$} & \textbf{Constant} & \textbf{64.98} & \textbf{74.42} & \textbf{69.70}\\
Adam $3\times10^{-3}$ & Constant & 64.52 & 74.10 & 69.31\\
Adam $10^{-3}$ & Cosine to $10^{-6}$ & 63.06 & 72.04 & 67.55\\
LARS & warmup + cosine & 52.78 & 63.02 & 57.90\\
\bottomrule
\end{tabular}
\end{table}

\subsection{SimCLR Implementation}
\label{app:simclr}

SimCLR uses a normalized temperature-scaled cross-entropy loss \citep{chen2020Simplea}. Two independently augmented views, with a minibatch of $B=512$ images, produce $2B$ projected feature vectors. The per example loss is:

\begin{equation}
\ell_i =
-\log
\frac{\exp(u_i^\top u_{j(i)}/\tau)}
{\sum_{\substack{k=1\\k\ne i}}^{2B}
\exp(u_i^\top u_k/\tau)}.
\label{eq:app_simclr_loss}
\end{equation}

Where $u_i$ is a projected vector of one view of an image and $j(i)$ the index of the other view of the same image. The overall loss is obtained by averaging the per example loss over all $2B$ views, with each view serving as an anchor. Each anchor has one positive and $2B-2$ negatives. 

We used temperature $\tau=0.5$, matching the shallow-network SimCLR implementation of \citet{patacchiola2020SelfSupervised} and the official CIFAR-10 configuration of SimCLR \citep{chen2020Simplea}. Training used Adam with constant learning rate $10^{-3}$, as was done in \citet{patacchiola2020SelfSupervised} and other SimCLR visual representational learning work \citep{robinson2021Contrastive}. The optimizer and learning rate were also consistent with our Barlow Twins experiments. The projector was a two-layer MLP, $h\rightarrow2048\rightarrow128$, where $h$ is the encoder feature dimension. Projector dimension justification is described in Appendix~\ref{app:simclr_hyper}.

SimCLR runs used the same Conv4 encoders, batch size, freezing procedure, and unweighted cosine similarity $k$NN evaluation as the Barlow Twins experiments.

\subsubsection{Image Augmentation Pipeline}

We followed the augmentation pipeline from the official CIFAR-10 implementation of SimCLR \citep{chen2020Simplea}. Our only adjustment is increasing the minimum crop size from 8\% to 20\%. This matched our Barlow Twins results and also generally improved terminal $k$NN accuracy (Table~\ref{tab:simclr_crop}). Each view underwent random resized cropping to $32\times32$, with crop-area fraction $[0.2,1]$, aspect ratio $[3/4,4/3]$, and bicubic interpolation with antialiasing. We then applied horizontal flipping with probability 0.5, color jitter with probability 0.8, and grayscale conversion with probability 0.2. Color-jitter magnitudes were $(0.4,0.4,0.4,0.1)$ for brightness, contrast, saturation, and hue. Gaussian blur and solarization were omitted. 

\begin{table}[t]
\centering
\small
\caption{Effect of minimum crop-area fraction on terminal test
$k$NN probe accuracy (\%) for Conv4 trained end-to-end to minimize SimCLR loss on CIFAR-10. Greedy layer-wise is included as reference.}
\label{tab:simclr_crop}
\begin{tabular}{@{}lrrrr@{}}
\toprule
& \multicolumn{2}{c}{End-to-end}
& \multicolumn{2}{c}{Greedy layer-wise} \\
\cmidrule(lr){2-3}\cmidrule(lr){4-5}
Width & 8\% crop & 20\% crop & 8\% crop & 20\% crop \\
\midrule
$0.25\times$ & 58.75 & 58.11 & 53.73 & 55.74 \\
$0.5\times$  & 63.80 & 64.00 & 61.00 & 60.90 \\
$1\times$    & 68.40 & 69.15 & 65.88 & 66.87 \\
$2\times$    & 72.41 & 72.84 & 69.37 & 70.59 \\
$4\times$    & 74.28 & 75.23 & 72.25 & 73.52 \\
$8\times$    & 75.57 & 75.77 & 73.95 & 74.05 \\
$16\times$   & 76.24 & 76.58 & 74.76 & 74.82 \\
$32\times$   & 76.40 & 76.24 & 75.00 & 75.92 \\
\midrule
Mean         & 70.73 & 70.99 & 68.24 & 69.05 \\
\bottomrule
\end{tabular}
\end{table}

\subsubsection{Hyperparameter Screens}
\label{app:simclr_hyper}

We selected the projector hidden dimension by performing a hyperparameter sweep with end-to-end Conv4 models at representative ($1\times$ and $8\times$) widths. We trained each candidate for 200 epochs with seed 0, batch size 512, temperature $\tau=0.5$, and Adam at a constant learning rate of $10^{-3}$. The projector output dimension was fixed at 128, as was done in the original SimCLR paper \citep{chen2020Simplea} and \citet{patacchiola2020SelfSupervised} while the hidden dimension varied over $\{128,256,512,1024,2048\}$. Each screen used a class-balanced split of the CIFAR-10 training set: 45,000 images for SSL training and the $k$NN memory bank, and 5,000 separate images for validation.

A hidden dimension of 2048 had the highest terminal validation accuracy at both widths (Table~\ref{tab:simclr_projector}). 

\begin{table}[htbp]
\centering
\small
\caption{Projector hidden-dimension screen for models trained end-to-end with SimCLR loss. Reported values are terminal validation
$k$NN probe accuracy on CIFAR-10.}
\label{tab:simclr_projector}
\begin{tabular}{@{}rrrr@{}}
\toprule
Hidden dimension & $1\times$ & $8\times$ & Mean\\
\midrule
128  & 63.28 & 70.16 & 66.72\\
256  & 64.02 & 70.68 & 67.35\\
512  & 64.40 & 71.10 & 67.75\\
1024 & 64.36 & 70.98 & 67.67\\
\textbf{2048} & \textbf{64.88} & \textbf{71.80} & \textbf{68.34}\\
\bottomrule
\end{tabular}
\end{table}

\subsection{Representational Geometry}
\label{app:representational_geometry}

We adopted the framework of \citet{wakhloo2026Neurala} to measure how greedy layer-wise and end-to-end training organize class-related information. We used all $N=10{,}000$ unaugmented CIFAR-10 test images with $K=10$ equally represented classes. Projectors were not used for this analysis. Class identity was used as the latent variable.

Images received the same input normalization used for $k$NN evaluation. At each layer, we extracted activations after convolution, BatchNorm, and ReLU. We then applied global average pooling. We L2-normalized each vector to match the features used for cosine $k$NN evaluation, then subtracted the mean vector across all test images. The mean was computed separately
for each model and layer over all 10,000 images. These vectors were stacked to form $X\in\mathbb{R}^{N\times D}$, where each row corresponds to an image and each column corresponds to one of the layer's $D$ channels. This $X$ matrix was used in subsequent calculations.

We constructed a one-hot matrix $Y\in\mathbb{R}^{N\times K}$, where $Y_{ik}=1$ if image $i$ belongs to class $k$ and zero otherwise. All images of a class had identical latent vectors. We centered each column by subtracting its mean. This ensured that the $\Psi$, $\Omega$, and $\Phi$ calculations below calculated covariances. Because the centered columns are linearly dependent, we could express them using nine orthonormal class contrasts and scaled these to unit variance. This allowed us to keep all class information while making the latent covariance matrix invertible, therefore allowing us to perform necessary future calculations. This produced the final latent matrix $Z$.

Following \citet{wakhloo2026Neurala}, we calculated
\begin{equation}
\mathbf{\Psi}=\frac{X^\top X}{N},
\qquad
\mathbf{\Omega}=\frac{Z^\top Z}{N},
\qquad
\mathbf{\Phi}=\frac{X^\top Z}{N},
\end{equation}
where $\mathbf{\Psi}$ is the representation covariance matrix,
$\mathbf{\Omega}$ is the latent covariance matrix, and
$\mathbf{\Phi}$ is their cross-covariance.

Signal--signal factorization (SSF) was calculated as
\begin{equation}
f=
\frac{
\left[\operatorname{Tr}
(\mathbf{\Phi}\mathbf{\Phi}^{\top})\right]^2
}{
\operatorname{Tr}(\mathbf{\Omega})\,
\operatorname{Tr}\!\left(
\mathbf{\Phi}^{\top}\mathbf{\Phi}
\mathbf{\Omega}^{-1}
\mathbf{\Phi}^{\top}\mathbf{\Phi}
\right)
}.
\label{eq:app_ssf}
\end{equation}

Here, $\operatorname{Tr}$ denotes the matrix trace.

Signal--noise factorization (SNF) was calculated as
\begin{equation}
s=
\frac{
\left[\operatorname{Tr}
(\mathbf{\Phi}\mathbf{\Phi}^{\top})\right]^2
}{
\operatorname{Tr}(\mathbf{\Omega})\,
\operatorname{Tr}\!\left[
\mathbf{\Phi}^{\top}
\left(
\mathbf{\Psi}-
\mathbf{\Phi}\mathbf{\Omega}^{-1}\mathbf{\Phi}^{\top}
\right)
\mathbf{\Phi}
\right]
}.
\label{eq:app_snf}
\end{equation}

To calculate the representational geometry, we used saved Conv4 encoders at width multipliers $\{0.25,0.5,1,2,4,8,16,32\}$. The geometry was calculated for both training procedures and all three seeds. Every encoder layer was measured and each metric was calculated separately for each seed.

The geometry over the course of training analysis (Fig.~\ref{fig:representation_geometry}c, d) used separate $32\times$ Conv4 runs with seed 1. Geometry was measured every 20 global epochs and at the end of each greedy stage. 1,000 epochs were run with 250 epochs per greedy layer, as was done in the main experiments.

\subsection{Reproducibility}
\label{app:reproducibility}

To support reproducibility, we used identical encoder-initializations and matched data-loading seeds for paired end-to-end and greedy runs. Primary experiments used three matched seeds with means and sample standard deviations reported. We additionally seeded Python, NumPy, PyTorch, and data-loader workers.

\end{document}